\documentclass[table]{style/style}

\usepackage{microtype}
\usepackage{hyperref}
\usepackage{url}
\usepackage{booktabs}
\usepackage{graphicx}
\usepackage{lineno}
\usepackage{enumitem}
\usepackage{listings} %
\usepackage{svg}

\definecolor{ darkblue}{rgb}{0, 0, 0.5}
\hypersetup{colorlinks=true, citecolor=darkblue, linkcolor=darkblue, urlcolor=darkblue}

\usepackage{amssymb}
\usepackage{multirow}
\usepackage{bigdelim}
\usepackage{todonotes}
\usepackage{longtable}
\usepackage{tabularray}
\usepackage{wrapfig}
\usepackage[most]{tcolorbox} 
\usepackage{url}
\usepackage{xspace}
\usepackage{svg}
\usepackage[absolute]{textpos} %

\usepackage{fdsymbol}   %

\usepackage[utf8]{inputenc} %
\usepackage[T1]{fontenc}    %
\usepackage{url}            %
\usepackage{booktabs}       %
\usepackage{amsfonts}       %
\usepackage{nicefrac}       %
\usepackage{microtype}      %
\usepackage[table,dvipsnames]{xcolor}         %
\usepackage{amsmath}
\usepackage[most]{tcolorbox}
\usepackage{csquotes}
\usepackage{wrapfig}
\usepackage{siunitx}
\usepackage{graphicx}
\usepackage{arydshln}
\usepackage{wrapfig}
\usepackage{enumitem}
\usepackage{soul} %

\usepackage{multirow}
\usepackage{xspace}
\usepackage{adjustbox}
\usepackage{pifont}
\usepackage{caption}
\usepackage{makecell}
\usepackage{subcaption}
\usepackage{bold-extra}
\usepackage{url}
\usepackage{float}
\usepackage{pgf-pie}

\usepackage{hyperref}
\crefname{figure}{Figure}{Figures}
\crefname{section}{Section}{Sections}
\crefname{equation}{Equation}{Equations}
\crefname{appendix}{Appendix}{Appendice}
\crefname{table}{Table}{Tables}

\definecolor{linkcolor}{RGB}{0, 0, 128}
\hypersetup{
     colorlinks   = true,
     citecolor    = linkcolor,
     linkcolor    = linkcolor,
     urlcolor     = linkcolor,
}
\usepackage{pifont}%
\usepackage{listings}

\setlist[itemize]{leftmargin=*,itemsep=0em,parsep=0.3em,topsep=0.3em}

\DeclareUnicodeCharacter{2212}{\ensuremath{-}}

\usepackage{tikz}

\usepackage{setspace}

\usepackage{nicematrix}
\newcolumntype{L}[1]{>{\raggedright\let\newline\\\arraybackslash\hspace{0pt}}m{#1}}
\newcolumntype{C}[1]{>{\centering\let\newline\\\arraybackslash\hspace{0pt}}m{#1}}
\newcolumntype{R}[1]{>{\raggedleft\let\newline\\\arraybackslash\hspace{0pt}}m{#1}}
\newcolumntype{P}[1]{>{\centering\let\newline\\\arraybackslash\hspace{0pt}}m{#1}}
\colorlet{lightgray}{White!30!lightgray}
\colorlet{lightblue}{White!70!MidnightBlue}

\title{\LARGE Context-aware Decoding Reduces Hallucination in Query-focused Summarization}

\authorOne{Zhichao Xu}
\affiliation{University of Utah}
\contribution[]{Corresponds to: \texttt{zhichao.xu@utah.edu}}

\abstract{
Query-focused summarization (QFS) aims to provide a summary of a single document/multi documents that can satisfy the information needs of a given query. It is useful for various real-world applications, such as abstractive snippet generation or more recent retrieval augmented generation (RAG). A prototypical QFS pipeline consists of a retriever (sparse or dense retrieval) and a generator (usually a large language model). However, applying large language models (LLM) potentially leads to hallucinations, especially when the evidence contradicts the prior belief of LLMs. There has been growing interest in developing new decoding methods to improve generation quality and reduce hallucination. In this work, we conduct a large-scale reproducibility study on one recently proposed decoding method\,---\,Context-aware Decoding (CAD).
In addition to replicating CAD's experiments on news summarization datasets, we include experiments on QFS datasets, and conduct more rigorous analysis on computational complexity and hyperparameter sensitivity. Experiments with eight different language models show that performance-wise, CAD improves QFS quality by (1) reducing factuality errors/hallucinations while (2) mostly retaining the match of lexical patterns, measured by ROUGE scores, while also at a cost of increased inference-time FLOPs and reduced decoding speed. The \href{https://github.com/zhichaoxu-shufe/context-aware-decoding-qfs}{code implementation} based on Huggingface Library is made available
}

\begin{document}

\maketitle

\section{Introduction}
Query-focused summarization (QFS) aims to provide a summary of a single document/multiple documents that can satisfy the information need of a given query~\citep{xu2023lightweight,marchionini2006exploratory}. 
It is useful for various real-world applications, such as abstractive snippet generation~\citep{chen2020abstractive} or more recent retrieval augmented generation (RAG)~\citep{lewis2020retrieval,gao2024retrievalaugmentedgenerationlargelanguage}. 
A prototypical QFS pipeline consists of a retriever to retrieve the documents to use as evidence, and a generator (usually based on large language models) to read the documents and generate natural language answers/summaries.

While the idea of combining large language models (LLM) and information retrieval seems tempting~\citep{metzler2021rethinking}, a few problems persist. 
For example, deploying generative models requires more inference FLOPs and specialized hardwares (GPUs or TPUs); and more importantly, the usage of LLMs in QFS/RAG pipeline also potentially leads to the hallucination problem where the generated summary contains information contradicting the source documents/evidence~\citep{shah2022situating,bender2021dangers}, which might hurt users' experience in information seeking processes and potentially escalate the spread of misinformation. 
As such, mainstream search engines (Google, Bing, etc.) still deploy extractive snippets~\citep{bast2014efficient}.

In parallel to training better language models to reduce hallucinations, there has been growing interest in developing decoding methods to allow the language models to better attend to evidence or to penalize degeneration, thus improving generation quality. For example, \citet{li-etal-2023-contrastive} propose contrastive decoding, leveraging the output probability distribution difference between an expert LM and an amateur LM to improve generation quality. 
DExperts~\citep{liu-etal-2021-dexperts} focus on reducing toxicity by contrasting the expert LM with toxicity LM.
\cite{xu2023lightweight} show that directly injecting knowledge constraints can potentially produce more faithful summarization.
A few works~\citep{shi2023trusting,van2022mutual} make use of pointwise mutual information to let the LM focus more on the input context to reduce hallucination. 

In this work, we focus on Context-aware Decoding~\citep[CAD,][]{shi2023trusting}. 
CAD leverages the idea of pointwise mutual information and proposes a product-of-experts enhancement to the vanilla decoding method to make the generation more conditioned on the input evidence. 
We conduct a large-scale reproducibility study with different choices of language models to study CAD's effectiveness in query-focused summarization, and include more rigorous analysis on computational complexity and hyperparameter sensitivity.
Extensive experiments show that CAD improves QFS quality by (1) reducing factual mistakes/hallucinations while (2) mostly retaining the match of lexical patterns, measured by ROUGE scores. 
On the other hand, CAD also introduces additional inference-time FLOPs and potentially slows down decoding speed.

\section{Background}
\label{section:background}
\paragraph{Context-aware Decoding.}
Consider a language model parameterized by $\theta$, a context $\mathbf{c}$, an input query $\mathbf{x}$, and we want to generate a piece of text $\mathbf{y}$. 
In QFS paradigm, $\mathbf{c}$ can be the source document, $\mathbf{x}$ and $\mathbf{y}$ will be the query and summary, respectively.
In the vanilla decoding scenario, each token of generation $\mathbf{y}$ is sampled autoregressively from the output probability distribution from language model $\theta$ conditioned on input query $\mathbf{x}$ and context $\mathbf{c}$:
\begin{equation}
    y_t \sim p_{\theta} (y_t | \mathbf{c}, \mathbf{x}, \mathbf{y}_{<t}) \propto \text{Softmax}\big( \, \text{logit}_{\theta} (y_t | \mathbf{c}, \mathbf{x}, \mathbf{y}_{<t}) / \tau \big)
\end{equation}
Here $\tau$ is a temperature hyperparameter to control the sharpness of the probability distribution over the candidate vocabulary $\mathcal{V}$.
Examples of the sampling strategy include greedy and Top-$p$ sampling~\citep{holtzman2019curious}.

As we are operating on a conditional generation problem, we hope the generation is more conditioned on input context $\mathbf{c}$. 
\cite{shi2023trusting} utilize the idea of pointwise mutual information (PMI).
We consider two events $\mathbf{c}$ and $y_t$:
\begin{equation}
    \mathbf{PMI} ( p_{\theta} (y_t ; \mathbf{c}, \mathbf{x}, \mathbf{y}_{<t}) ) = \log \Big( \frac{ p_{\theta}(y_t | \mathbf{c}, \mathbf{x}, \mathbf{y}_{<t})} { p_{\theta}(y_t | \mathbf{x}, \mathbf{y}_{<t})} \Big)
\end{equation}
The interpretation of the above equation is we would like to measure the association of event $y_t$ (predicting a specific token) and event $\mathbf{c}$ (the presence of context $\mathbf{c}$).
$p_{\theta}(y_t | \mathbf{x}, \mathbf{y}_{<t})$ is the prior probability, representing the model's prior belief from its parameters $\theta$ when no presence of context $\mathbf{c}$, whereas $p_{\theta}(y_t | \mathbf{c}, \mathbf{x}, \mathbf{y}_{<t})$ is the posterior probability with $\mathbf{c}$.
To leverage PMI, \citet{shi2023trusting} propose to multiply PMI with the original probability distribution:
\begin{equation}
    y_t \sim \tilde{p}_{\theta} (y_t | \mathbf{c}, \mathbf{x}, \mathbf{y}_{<t}) \propto p_{\theta} (y_t | \mathbf{c}, \mathbf{x}, \mathbf{y}_{<t}) \Big( \frac{ p_{\theta}(y_t | \mathbf{c}, \mathbf{x}, \mathbf{y}_{<t})} { p_{\theta}(y_t | \mathbf{x}, \mathbf{y}_{<t})} \Big)^{\alpha}
\end{equation}
Here $\alpha$ is a hyperparameter to control the PMI weighting.
We still need to normalize the above equation to acquire a valid probability distribution, rearranging the terms yields
\begin{equation}
    y_t \sim \text{Softmax} \big[ \big( (1+\alpha) \, \text{logit}_{\theta}(y_t | \mathbf{c}, \mathbf{x}, \mathbf{y}_{<t}) - \alpha \,\text{logit}_{\theta}(y_t | \mathbf{x}, \mathbf{y}_{<t}) \big) / \tau \big]
\end{equation}
Here larger $\alpha$ yields more weight on the context $\mathbf{c}$ and $\alpha=0$ reduces to vanilla decoding. 
\cite{shi2023trusting} recommend setting $\alpha=0.5$ for summarization tasks and we also examine this choice in Section~\ref{section:experiments}.

\paragraph{Analysis of Computational Cost.}
We follow the notations used in~\cite{kaplan2020scaling}. Let us consider the Transformer architecture with following hyperparameters: number of layers $n_{\text{layer}}$, dimension of the embedding layer and intermediate representation $d_{\text{model}}$, dimension of the intermediate feedforward layer $d_{\text{ff}}$, dimension of the attention output $d_{\text{attn}}$ and number of attention heads $n_{\text{heads}}$; and denote the number of tokens in the input sequence as $n_{\text{input}}$.
The model size $N$ is the number of \emph{non-embedding} parameters of the model, $N\approx |\theta|$:
\begin{equation}
    N \approx 2\, d_{\text{model}}\, n_{\text{layer}} (2\, d_{\text{attn}}+d_{\text{ff}}) = 12\, n_{\text{layer}} \,d_{\text{model}}^2  \,\,\,\, \text{with} \,\, d_{\text{attn}}=d_{\text{ff}}/4 = d_{\text{model}}
\end{equation}
The total FLOPs per token in one forward pass is approximately $C_{\text{forward}}=2N+2 n_{\text{layer}} \, n_{\text{input}} \, d_{\text{attn}}$ (see Section 2.1 in~\cite{kaplan2020scaling} for details). 
In LLMs, $d_{\text{model}} \gg n_{\text{input}}/12$, we can roughly ignore the second term and use $C_{\text{forward}}\approx 2N$.

With the above analysis, we know that performing one forward pass of CAD yields in total $(2\,|\mathbf{y}_{<t}|+2\,|\mathbf{x}|+|\mathbf{c}|) \cdot C_\text{forward}$ FLOPs compared to $(|\mathbf{y}_{<t}|+|\mathbf{x}| +|\mathbf{c}|) \cdot C_\text{forward}$ from vanilla decoding.

\section{Experiments}
\label{section:experiments}

We carry out experiments on two QFS datasets and two news summarization datasets, with different choices of backbone language models, including pre-trained language models and instruction finetuned language models.

\subsection{Prompting Templates}
For QFS task, we treat the source document as context $\mathbf{c}$, and treat the original query as query $\mathbf{x}$ and prompt the model to generate summarization $\mathbf{y}$. 
For the summarization task, the source document is used as context $\mathbf{c}$, and we manually append a query $\mathbf{x}$ in the shape of \texttt{[summarize the following document:]}. The detailed prompting templates can be found in Appendix~\ref{subsec:templates}.

\subsection{Experiment Setup}
\paragraph{Datasets.} We benchmark CAD's performance on two widely adopted QFS datasets. Debatepedia dataset is collected by~\cite{nema-etal-2017-diversity} from 663 debates of 53 diverse categories in an encyclopedia of debates and consists of 12K/0.7K/1.0K query-document summarization triplets $(\emph{query}, \emph{document}, \emph{summarization})$.
PubMedQA~\citep{jin-etal-2019-pubmedqa} is a long-form abstractive question-answering dataset from the biomedical domain with the contexts available.
We also use two news summarization datasets from the original CAD paper: CNN Dailymail~\citep{see-etal-2017-get} and XSUM~\citep{narayan-etal-2018-dont} for reproducibility purposes.

\paragraph{Language Models.} 
We study language models of different architecture choices (encoder-decoder models vs decoder-only models) and training paradigms (pre-trained vs instruction finetuned).
\begin{itemize}
    \item \texttt{LLaMA-7B}~\citep{touvron2023llama} is a decoder-only language model pre-trained with causal language modeling objective from FAIR.
    \item \texttt{Llama-2-7B}~\citep{touvron2023llamatwo} is the second iteration of \texttt{LLaMA} model by better pre-training corpus and minor architectural improvements. 
    \item \texttt{MPT-7B}~\citep{MosaicML2023Introducing} is a decoder-only language model pre-trained with causal language modeling objective from MosaicML
    \item \texttt{MPT-7B-instruct} is based on \texttt{MPT-7B} and further instruction finetuned with an instruction finetuning dataset \texttt{mosaicml/instruct-v1}.
    \item \texttt{Mistral-7B-v0.1}~\citep{jiang2023mistral} is a decoder-only language model pre-trained with causal language modeling objective from Mistral AI.
    \item \texttt{Mistral-7B-instruct-v0.1} is based on \texttt{Mistral-7B-v0.1} and further instruction finetuned with an internal dataset.
    \item \texttt{Flan-T5-xl}~\citep{chung2022scaling} (3B) is an encoder-decoder model from T5 family~\cite{raffel2020exploring}; it is pre-trained with span-infilling objective and further adapted with the causal language modeling objective. It is instruction finetuned on the Flan collection.
    \item \texttt{Flan-T5-xxl}~\citep{chung2022scaling} (11B) undergoes same training procedure as \texttt{Flan-T5-xl} and demonstrates strong performance on natural language understanding tasks. 
\end{itemize}
We acquire weights of all listed models from Huggingface library\footnote{\url{https://huggingface.co/models}}.

\paragraph{Evaluation.}
Similar to~\cite{shi2023trusting}, we evaluate the generation quality with ROUGE F1~\citep{lin-2004-rouge}, BERTScore-precision~\citep{zhang2019bertscore} and FactKB~\citep{feng2023factkb}. 
ROUGE measures the match of lexical patterns while BERTScore measures the semantic similarity between the generation and the reference.
FactKB is a model-based evaluation metric that classifies whether the generation is faithful to the document.
For BERTScore we use \texttt{roberta-large-mnli} as it shows a higher correlation with human judgments. 
For FactKB we use the original classifier trained by the authors~\citep{feng2023factkb}.

\paragraph{Hyperparameters}
Our experimental settings follow~\cite{shi2023trusting} to a large extent. 
For decoding, we set temperature=1, use a combination of Top-$k$ and Top-$p$ sampling, and vary $\alpha$ to study the effectiveness of CAD. 
For the main results reported in Section~\ref{sec:results}, we use $\alpha=0.3$ and use 0.15, 0.5 as an ablation study.
Detailed choice of hyperparameters can be found in Appendix~\ref{subsec:hyperparameters}.

\section{Results and Analysis}
\label{sec:results}

\begin{table*}[!t]
\centering
\caption{Vanilla decoding vs CAD with news summarization datasets. }
\resizebox{0.9\textwidth}{!}{
\begin{tabular}{llcrrrrr}
\toprule
\textbf{Datasets}  & \textbf{Model} & \textbf{Decoding} & \textbf{ROUGE-1} & \textbf{ROUGE-2} & \textbf{ROUGE-L} & \textbf{BERTScore-P} & \textbf{FactKB}\\ 
\hline
\multirow{16}{*}{CNN Dailymail} & \multirow{2}{*}{\texttt{LLaMA-7B}} & Vanilla & 27.6 & 8.7 & 24.5 & 73.0 & 85.8 \\
& & CAD & 30.4 & 10.8 & 27.1 & 73.5 & 90.9 \\ \cline{2-8}
& \multirow{2}{*}{\texttt{Llama-2-7B}} & Vanilla & 26.8 & 8.6 & 23.9 & 72.3 & 80.6 \\
& & CAD & 29.3 & 10.3 & 26.2 & 73.1 & 84.9 \\ \cline{2-8}
& \multirow{2}{*}{\texttt{MPT-7B}} & Vanilla & 28.9 & 9.8 & 25.7 & 73.0 & 86.2 \\
& & CAD & 30.5 & 11.1 & 27.0 & 73.8 & 90.4 \\ \cline{2-8}
& \multirow{2}{*}{\texttt{MPT-7B-Instruct}} & Vanilla & 31.0 & 10.6 & 27.4 & 74.2 & 89.2 \\
& & CAD & 32.2 & 11.9 & 28.5 & 74.9 & 94.0 \\ \cline{2-8}
& \multirow{2}{*}{\texttt{Mistral-7B}} & Vanilla & 28.1 & 9.3 & 24.9 & 72.8 & 78.5 \\
& & CAD & 29.3 & 10.6 & 25.9 & 73.7 & 84.0 \\ \cline{2-8}
& \multirow{2}{*}{\texttt{Mistral-7B-Instruct}} & Vanilla & 34.2 & 11.4 & 30.5 & 75.4 & 93.1 \\
& & CAD & 34.6 & 13.3 & 30.9 & 75.5 & 95.6 \\ \cline{2-8}
& \multirow{2}{*}{\texttt{Flan-T5-xl}} & Vanilla & 29.0 & 10.0 & 25.6 & 75.6 & 86.0 \\
& & CAD & 30.8 & 11.6 & 27.5 & 75.9 & 85.6 \\ \cline{2-8}
& \multirow{2}{*}{\texttt{Flan-T5-xxl}} & Vanilla & 27.0 & 8.4 & 24.0 & 74.5 & 82.3 \\
& & CAD & 28.8 & 10.0 & 25.8 & 75.1 & 81.8 \\
\midrule
\multirow{16}{*}{XSUM} & \multirow{2}{*}{\texttt{LLaMA-7B}} & Vanilla & 19.7 & 3.9 & 15.5 & 71.5 & 38.9 \\
& & CAD & 19.9 & 3.8 & 15.6 & 71.8 & 53.8 \\ \cline{2-8}
& \multirow{2}{*}{\texttt{Llama-2-7B}} & Vanilla & 19.5 & 3.9 & 15.3 & 71.5 & 47.2 \\
& & CAD & 20.0 & 4.0 & 15.6 & 71.7 & 58.1 \\ \cline{2-8}
& \multirow{2}{*}{\texttt{MPT-7B}} & Vanilla & 19.5 & 3.7 & 15.2 & 71.6 & 56.2 \\
& & CAD & 20.0 & 3.8 & 15.5 & 72.3 & 63.6 \\ \cline{2-8}
& \multirow{2}{*}{\texttt{MPT-7B-Instruct}} & Vanilla & 22.9 & 5.3 & 17.7 & 73.4 & 51.5 \\
& & CAD & 22.3 & 5.1 & 17.2 & 76.1 & 62.6 \\ \cline{2-8}
& \multirow{2}{*}{\texttt{Mistral-7B}} & Vanilla & 23.2 & 5.9 & 18.0 & 72.6 & 43.0 \\
& & CAD & 22.5 & 5.9 & 17.6 & 73.3 & 51.8 \\ \cline{2-8}
& \multirow{2}{*}{\texttt{Mistral-7B-Instruct}} & Vanilla & 25.8 & 6.7 & 20.0 & 74.2 & 55.7 \\
& & CAD & 25.2 & 6.5 & 19.5 & 73.9 & 63.4 \\ \cline{2-8}
& \multirow{2}{*}{\texttt{Flan-T5-xl}} & Vanilla & 34.8 & 12.8 & 27.5 & 80.4 & 53.3 \\
& & CAD & 35.8 & 13.8 & 28.5 & 80.8 & 55.6 \\ \cline{2-8}
& \multirow{2}{*}{\texttt{Flan-T5-xxl}} & Vanilla & 36.0 & 13.9 & 28.2 & 80.8 & 54.8 \\
& & CAD & 36.7 & 14.7 & 29.3 & 81.1 & 54.9 \\

\bottomrule
\end{tabular}
}
\label{tab:summarization_main}
\end{table*}

\begin{table*}[!t]
\centering
\caption{Vanilla decoding vs CAD with QFS datasets.}
\resizebox{0.9\textwidth}{!}{
\begin{tabular}{llcrrrrr}
\toprule
\textbf{Datasets}  & \textbf{Model} & \textbf{Decoding} & \textbf{ROUGE-1} & \textbf{ROUGE-2} & \textbf{ROUGE-L} & \textbf{BERTScore-P} & \textbf{FactKB}\\ 
\hline
\multirow{16}{*}{Dbpedia} & \multirow{2}{*}{\texttt{LLaMA-7B}} & Vanilla & 13.3 & 2.9 & 11.8 & 68.8 & 29.6 \\
& & CAD & 13.4 & 3.1 & 12.0 & 68.9 & 36.1 \\ \cline{2-8}
& \multirow{2}{*}{\texttt{Llama-2-7B}} & Vanilla & 12.0 & 2.5 & 10.8 & 68.6 & 27.3 \\
& & CAD & 12.7 & 2.7 & 11.2 & 68.4 & 34.0 \\ \cline{2-8}
& \multirow{2}{*}{\texttt{MPT-7B}} & Vanilla & 13.1 & 2.7 & 11.6 & 70.0 & 33.4 \\
& & CAD & 13.3 & 3.0 & 12.0 & 70.4 & 37.9 \\ \cline{2-8}
& \multirow{2}{*}{\texttt{MPT-7B-Instruct}} & Vanilla & 14.8 & 3.7 & 13.1 & 71.5 & 28.1 \\
& & CAD & 14.5 & 3.4 & 12.8 & 72.2 & 31.3 \\ \cline{2-8}
& \multirow{2}{*}{\texttt{Mistral-7B}} & Vanilla & 13.8 & 3.2 & 12.2 & 69.4 & 28.8 \\
& & CAD & 13.8 & 3.0 & 12.3 & 69.1 & 36.9 \\ \cline{2-8}
& \multirow{2}{*}{\texttt{Mistral-7B-Instruct}} & Vanilla & 18.2 & 5.3 & 15.9 & 71.6 & 42.6 \\
& & CAD & 17.1 & 4.7 & 15.1 & 71.1 & 41.6 \\ \cline{2-8}
& \multirow{2}{*}{\texttt{Flan-T5-xl}} & Vanilla & 18.0 & 5.4 & 16.2 & 74.4 & 37.5 \\
& & CAD & 16.2 & 4.5 & 14.5 & 73.7 & 46.3 \\ \cline{2-8}
& \multirow{2}{*}{\texttt{Flan-T5-xxl}} & Vanilla & 20.6 & 6.8 & 18.6 & 75.0 & 33.9 \\
& & CAD & 16.2 & 4.4 & 14.5 & 73.5 & 43.0 \\
\midrule
\multirow{16}{*}{PubMedQA} & \multirow{2}{*}{\texttt{LLaMA-7B}} & Vanilla & 25.5 & 7.6 & 21.3 & 71.2 & 79.5 \\
& & CAD & 26.0 & 8.2 & 21.6 & 71.3 & 86.4 \\ \cline{2-8}
& \multirow{2}{*}{\texttt{Llama-2-7B}} & Vanilla & 27.7 & 9.0 & 23.0 & 72.0 & 87.4 \\
& & CAD & 28.6 & 9.5 & 23.6 & 72.1 & 90.8 \\ \cline{2-8}
& \multirow{2}{*}{\texttt{MPT-7B}} & Vanilla & 27.8 & 8.7 & 23.0 & 72.0 & 92.4 \\
& & CAD & 27.4 & 8.5 & 22.8 & 71.7 & 96.3 \\ \cline{2-8}
& \multirow{2}{*}{\texttt{MPT-7B-Instruct}} & Vanilla & 29.0 & 9.7 & 24.0 & 72.9 & 96.4 \\
& & CAD & 27.6 & 8.7 & 22.6 & 72.2 & 97.3 \\ \cline{2-8}
& \multirow{2}{*}{\texttt{Mistral-7B}} & Vanilla & 25.0 & 8.0 & 20.8 & 71.7 & 67.9 \\
& & CAD & 24.4 & 7.6 & 20.1 & 71.5 & 75.7 \\ \cline{2-8}
& \multirow{2}{*}{\texttt{Mistral-7B-Instruct}} & Vanilla & 31.7 & 11.9 & 26.4 & 74.3 & 97.5 \\
& & CAD & 31.2 & 11.6 & 26.0 & 73.9 & 97.5 \\ \cline{2-8}
& \multirow{2}{*}{\texttt{Flan-T5-xl}} & Vanilla & 32.3 & 11.4 & 26.6 & 76.1 & 95.7 \\
& & CAD & 30.5 & 10.1 & 24.9 & 75.3 & 95.2 \\ \cline{2-8}
& \multirow{2}{*}{\texttt{Flan-T5-xxl}} & Vanilla & 33.1 & 11.6 & 27.0 & 76.3 & 95.6 \\
& & CAD & 31.3 & 10.6 & 25.6 & 75.6 & 94.3 \\

\bottomrule
\end{tabular}
}
\label{tab:qfs_main}
\end{table*}

We try to study the following research questions:
\begin{itemize}
    \item RQ1: Performance-wise, can CAD (1) improve news summarization performance as discussed by~\cite{shi2023trusting}? and (2) reduce hallucination/factuality errors in QFS?
    \item RQ2: What is a good choice of hyperparameter $\alpha$?
    \item RQ3: What is the computational trade-off?
\end{itemize}
\paragraph{Quality of Generation.}
We present the results on QFS datasets and news summarization datasets in Table~\ref{tab:qfs_main} and Table~\ref{tab:summarization_main}, respectively.
On news summarization datasets, CAD achieves improved ROUGE scores with all eight models on CNN Dailymail dataset, and improves with six models on XSUM datasets. 
In terms of factuality errors, CAD improves FactKB scores on six out of eight models on CNN Dailymail dataset and eight out of eight models on XSUM dataset. 
These results support the finding in~\cite{shi2023trusting} that CAD can improve news summarization performance.

In terms of results on QFS datasets, we notice that while CAD can reduce hallucination/factuality errors as measured by FactKB score, the ROUGE score is not improved on Dbpedia dataset and slightly drops on PubMedQA dataset.
We hypothesize that the different performance patterns might be because the references in two QFS datasets are highly abstractive, and myopically conditioning on the input context may not achieve an improved ROUGE score.

As for performance between different language models, a clear observation is that instruction finetuned models (\texttt{MPT-7B-Instruct} and \texttt{Mistral-7B-Instruct-v0.1}) outperform their pre-trained counterparts. 
On the relatively challenging PubMedQA dataset, \texttt{Mistral-7B-Instruct-v0.1} achieves comparable performance to \texttt{Flan-T5-xxl}, despite \texttt{Flan-T5-xxl} being trained on more academic style NLP tasks. 

\paragraph{Choice of hyperparameter $\alpha$.}
We show a comparison between different choices of hyperparameter $\alpha$ against model performance in Table~\ref{tab:hyperparameter}. 
From QFS results in Table~\ref{tab:hyperparameter} we notice that there is a trade-off between FactKB score and ROUGE score as $\alpha$ increases, whereas on the two news summarization datasets, the ROUGE score and FactKB score are positively correlated. 
We hypothesize that the different trends are due to the design choices in building the datasets.
On the news summarization datasets, setting $\alpha=0.3$ yields an overall better performance.

\begin{table*}[!t]
\centering
\captionsetup{width=0.8\columnwidth}
\caption{Performance comparison between difference choices of hyperparameter $\alpha$, recall that $\alpha=0$ reduces to vanilla decoding. The reported numbers are averaged over eight models.}
\resizebox{0.8\columnwidth}{!}{
\begin{tabular}{llrrrrr}
\toprule
\textbf{Datasets}  & $\boldsymbol{\alpha}$ \textbf{value} & \textbf{ROUGE-1} & \textbf{ROUGE-2} & \textbf{ROUGE-L} & \textbf{BERTScore-P} & \textbf{FactKB}\\ 
\hline
\multirow{4}{*}{Dbpedia} & $\alpha$=0.0 & 15.5 & 4.1 & 13.8 & 71.2 & 32.7 \\
& $\alpha$=0.15 & 14.9 & 3.8 & 13.3 & 71.1 & 35.1 \\
& $\alpha$=0.3 & 14.7 & 3.6 & 13.1 & 70.1 & 38.1 \\
& $\alpha$=0.5 & 14.5 & 3.6 & 13.0 & 70.9 & 38.7 \\
\midrule
\multirow{4}{*}{PubMedQA} & $\alpha$=0.0 & 29.1 & 9.7 & 24.0 & 73.3 & 89.1 \\
& $\alpha$=0.15 & 28.7 & 9.6 & 23.8 & 73.2 & 90.6 \\
& $\alpha$=0.3 & 28.4 & 9.4 & 23.4 & 73.0 & 91.7 \\
& $\alpha$=0.5 & 27.9 & 9.1 & 23.1 & 72.7 & 93.3 \\
\midrule
\multirow{4}{*}{CNN Dailymail} & $\alpha$=0.0 & 29.1 & 9.6 & 25.8 & 73.9 & 85.2 \\
& $\alpha$=0.15 & 30.4 & 10.9 & 27.1 & 74.3 & 87.2 \\
& $\alpha$=0.3 & 30.7 & 11.2 & 27.4 & 74.4 & 88.4 \\
& $\alpha$=0.5 & 30.7 & 11.2 & 27.3 & 74.4 & 87.7 \\
\midrule
\multirow{4}{*}{XSUM} & $\alpha$=0.0 & 25.2 & 7.0 & 19.7 & 74.5 & 50.1 \\
& $\alpha$=0.15 & 25.4 & 7.2 & 19.8 & 74.8 & 55.7 \\
& $\alpha$=0.3 & 25.3 & 7.2 & 19.9 & 74.9 & 58.0 \\
& $\alpha$=0.5 & 24.8 & 6.9 & 19.4 & 74.9 & 58.1 \\
\bottomrule
\end{tabular}
}
\label{tab:hyperparameter}
\end{table*}

\paragraph{Reporting Decoding Speed.}
Previously we have analyzed the additional complexity in terms of FLOPs of CAD. 
Now we take a look at the difference in decoding speed on a real-world dataset.
The experiment is conducted on a server with a single Nvidia RTX 3090 GPU, CUDA version 12.1, and PyTorch version 2.1.0.
We implement two forward passes, first with $p_{\theta}(y_t|\mathbf{c}, \, \mathbf{x}, \, \mathbf{y}_{<t})$, and second with $p_{\theta}(y_t|\mathbf{x},\, \mathbf{y}_{<t})$; another option is packing two computation into one batch which will increase CUDA memory usage.

On CNN Dailymail dataset, vanilla decoding with FP16 and batch size=1 leads to a decoding speed of 0.042 second/token, while the decoding speed of CAD with FP16 and batch size=1 is 0.077 second/token with two forward passes. 
This suggests while CAD can improve generation quality, it also slows down decoding speed or requires more CUDA memory.

\section{Related Work}
\label{sec:related}

\paragraph{Search as an Information-seeking Process.}
\cite{marchionini2006exploratory} categorize search into three two different information seeking processes\,---\,\emph{Information Lookup} and \emph{Exploratory Search}.
Information lookup refers to the user looking for simple facts or asking factoid questions, while exploratory search/search to learn often involves multiple iterations.
Different approaches have been explored to reduce users' cognitive loads in the information-seeking process. 
For example, Perplexity AI\footnote{\url{https://www.perplexity.ai/}} transfers web search queries into QA-style questions, and cites sources in the answers to improve verifiability.
Snippets~\citep{bast2014efficient} provide short previews for each webpage to allow users to locate relevant information without examining the webpage.
Counterfactual explanations~\citep{xu2024counterfactual} suggest counterfactual queries as explanations and demonstrate search sessions' effectiveness can be improved.

\paragraph{Hallucination in Natural Language Generation.}
\citet{ji2023survey} present a survey of hallucination in natural language generation. 
There are mainly two types of hallucinations, namely intrinsic hallucination and extrinsic hallucination. 
Intrinsic hallucination refers to the generated output that contradicts the source content, whereas extrinsic hallucination refers to the generated output that cannot be verified from the source content. 
In retrieval augmented generation, we mainly focus on intrinsic hallucination.
Pre-trained language models (not instruction finetuned) are more prone to hallucination, as the gigantic pre-training corpus may contain untruthful texts and the teacher forcing training procedure lets the models pick up this untruthful information.
\citet{shuster-etal-2021-retrieval-augmentation} show that retrieval augmentation can reduce hallucinations in downstream tasks.
\citet{tian2023fine} show that language models can be trained to reduce hallucination via reinforcement learning techniques.
\citet{varshney2023stitch} propose to use lm logits to detect hallucination in the generation process and mitigate it by retrieving evidence and rewriting.

\paragraph{Decoding Methods to Improve Generation Quality.}
\citet{li-etal-2023-contrastive} propose Contrastive Decoding (CD) to improve generation quality. The idea is to contrast the stronger expert model's prediction with a smaller amateur model. 
A recent work~\citep{o2023contrastive} shows that CD can also improve the reasoning ability of the language models on multiple reasoning benchmark datasets.
Based on a similar idea, \citet{liu-etal-2021-dexperts} show that by contrasting to a toxic model (language model trained with toxicity contents), the toxicity of the expert language model's generations can be reduced.
A different flavor of controlled generation aims to let the model generate specific tokens/phrases. 
\citet{lu-etal-2021-neurologic} show that such a goal can be achieved by utilizing specific data structure and interpolation of lm logits.

\section{Conclusion and Limitations}
In this work we carry out a reproducibility study of a recently proposed decoding method\,---\,Context-aware Decoding (CAD). 
Experiments on query-focused summarization datasets and news summarization datasets show that CAD can improve QFS and news summarization quality by reducing factual mistakes/hallucinations while mostly retaining the match of lexical patterns, but also with additional inference-time FLOPs and reduced decoding speed.
Due to the limited bandwidth and resources, we only experiment with language models no larger than 11B (\texttt{Flan-T5-xxl}), larger language models may exhibit different performance patterns.

\bibliographystyle{plainnat}
\bibliography{references}

\section{Appendix}
\label{sec:appendix}

\subsection{Prompting Templates}
\label{subsec:templates}
We report the prompting templates in Table~\ref{tab:appendix_templates}.

\begin{table*}[!h]
\centering
\caption{Prompting Templates}
\vspace{-5pt}
\resizebox{\columnwidth}{!}{
\begin{tabular}{llll}
\toprule
\textbf{Datasets} & \textbf{Model} & \textbf{Decoding} & \textbf{Templates} \\ 
\midrule
\multirow{4}{*}{Dbpedia} & Decoder-only & Vanilla &
\begin{tabular}[c]{@{}l@{}l} \texttt{Question: \{query\}. Document: \{document\}.} \\ \texttt{According to the Document, the one-sentence answer to the Question is:} \end{tabular}\\
\cline{2-4}

 & Decoder-only & CAD &
\begin{tabular}[c]{@{}l@{}l} \texttt{Question: \{query\}. Document: \{None\}.} \\ \texttt{According to the Document, the one-sentence answer to the Question is:} \end{tabular}\\
\cline{2-4}

& Encoder-Decoder & Vanilla & \begin{tabular}[c]{@{}l@{}l} \texttt{Question: \{query\}. Document: \{document\}.} \\ \texttt{According to the Document, the one-sentence answer to the Question is:} \end{tabular}\\
\cline{2-4}
& Encoder-Decoder & CAD & \begin{tabular}[c]{@{}l@{}l} \texttt{Question: \{query\}. Document: \{None\}.} \\ \texttt{According to the Document, the one-sentence answer to the Question is:} \end{tabular}\\
\midrule

\multirow{4}{*}{PubMedQA} & Decoder-only & Vanilla &
\begin{tabular}[c]{@{}l@{}l} \texttt{Question: \{query\}. Document: \{document\}.} \\ \texttt{According to the Document, the detailed answer to the Question is:} \end{tabular}\\
\cline{2-4}

& Decoder-only & CAD &
\begin{tabular}[c]{@{}l@{}l} \texttt{Question: \{query\}. Document: \{document\}.} \\ \texttt{According to the Document, the detailed answer to the Question is:} \end{tabular}\\
\cline{2-4}

& Encoder-Decoder & Vanilla & \begin{tabular}[c]{@{}l@{}l} \texttt{Question: \{query\}. Document: \{document\}.} \\ \texttt{According to the Document, the detailed answer to the Question is:} \end{tabular}\\
\cline{2-4}

& Encoder-Decoder & CAD & \begin{tabular}[c]{@{}l@{}l} \texttt{Question: \{query\}. Document: \{None\}.} \\ \texttt{According to the Document, the detailed answer to the Question is:} \end{tabular}\\
\midrule

\multirow{4}{*}{CNN Dailymail} & Decoder-only & Vanilla &
\begin{tabular}[c]{@{}l@{}l} \texttt{News article: \{document\}.} \\ \texttt{Summary of the above news article:} \end{tabular}\\
\cline{2-4}

& Decoder-only & CAD &
\begin{tabular}[c]{@{}l@{}l} \texttt{News article: \{None\}.} \\ \texttt{Summary of the above news article:} \end{tabular}\\
\cline{2-4}

& Encoder-Decoder & Vanilla & \begin{tabular}[c]{@{}l@{}l} \texttt{Summarize the following article in one or two sentences.} \\ \texttt{\{document\}:} \end{tabular}\\
\cline{2-4}

& Encoder-Decoder & CAD & \begin{tabular}[c]{@{}l@{}l} \texttt{Summarize the following article in one or two sentences.} \\ \texttt{\{None\}:} \end{tabular}\\
\midrule

\multirow{4}{*}{XSUM} & Decoder-only & Vanilla &
\begin{tabular}[c]{@{}l@{}l} \texttt{News article: \{document\}.} \\ \texttt{Summary of the above news article:} \end{tabular}\\
\cline{2-4}

 & Decoder-only & CAD &
\begin{tabular}[c]{@{}l@{}l} \texttt{News article: \{None\}.} \\ \texttt{Summary of the above news article:} \end{tabular}\\
\cline{2-4}

& Encoder-Decoder & Vanilla & \begin{tabular}[c]{@{}l@{}l} \texttt{Summarize the following article in one or two sentences.} \\ \texttt{\{document\}:} \end{tabular}\\
\cline{2-4}

& Encoder-Decoder & CAD & \begin{tabular}[c]{@{}l@{}l} \texttt{Summarize the following article in one or two sentences.} \\ \texttt{\{None\}:} \end{tabular}\\

\bottomrule
\end{tabular}
}
\label{tab:appendix_templates}
\end{table*}

\subsection{Hyperparameters}
\label{subsec:hyperparameters}
We report the hyperparameters in Table~\ref{tab:appendix_hyperaprameters}.
\begin{table*}[!h]
\centering
\caption{Decoding Hyperparameters}
\vspace{-5pt}
\resizebox{\columnwidth}{!}{
\begin{tabular}{lrrrrrrr}
\toprule
\textbf{Dataset} & \textbf{Top-k} & \textbf{Top-p} & \textbf{num\_beams} & \textbf{repetition\_penalty} & \textbf{temperature} & \textbf{min\_new\_tokens} & \textbf{max\_new\_tokens} \\
\midrule
Dbpedia & 50 & 0.9 & 1 & 1.0 & 1.0 & 5 & 30 \\
PubMedQA & 50 & 0.9 & 1 & 1.0 & 1.0 & 40 & 100 \\
CNN Dailymail & 50 & 0.9 & 1 & 1.0 & 1.0 & 30 & 70 \\
XSUM & 50 & 0.9 & 1 & 1.0 & 1.0 & 20 & 50 \\

\bottomrule
\end{tabular}
}
\label{tab:appendix_hyperaprameters}
\end{table*}

\end{document}